\documentclass[preprint,12pt]{elsarticle}

\usepackage{blindtext}
\usepackage{amssymb}
\usepackage{amsmath}
\usepackage{tikz}
\usepackage{tikz-cd}
\usepackage{bm}
\usepackage{soul}
\usepackage{sectsty}

\usepackage{hyperref}
\usepackage{cleveref}

\begin{document}

\begin{frontmatter}

\title{Drug Release Modeling using Physics-Informed Neural Networks}

\author[inst1,inst2]{Daanish Aleem Qureshi}
\author[inst1]{Khemraj Shukla}
\author[inst2,inst3]{Vikas Srivastava\corref{cor1}}
\cortext[cor1]{Corresponding author: Vikas\_Srivastava@brown.edu}
\address[inst1]{Division of Applied Mathematics, Brown University}
\address[inst2]{School of Engineering, Brown University}
\address[inst3]{Institute for Biology, Engineering and Medicine, Brown University}


\begin{abstract}
Accurate modeling of drug release is essential for designing and developing controlled-release systems. Classical models (Fick, Higuchi, Peppas) rely on simplifying assumptions that limit their accuracy in complex geometries and release mechanisms. Here, we propose a novel approach using Physics-Informed Neural Networks (PINNs) and Bayesian PINNs (BPINNs) for predicting release from planar, 1D-wrinkled, and 2D-crumpled films. This approach uniquely integrates Fick’s diffusion law with limited experimental data to enable accurate long-term predictions from short-term measurements, and is systematically benchmarked against classical drug release models.  We embedded Fick’s second law into PINN as loss with 10,000 Latin-hypercube collocation points and utilized previously published experimental datasets to assess drug release performance through mean absolute error (MAE) and root mean square error (RMSE), considering noisy conditions and limited-data scenarios. Our approach reduced mean error by up to 40\% relative to classical baselines across all film types. The PINN formulation achieved RMSE $<0.05$ utilizing only the first 6\% of the release time data (reducing 94\% of release time required for the experiments) for the planar film. For wrinkled and crumpled films, the PINN reached RMSE $<0.05$  in 33\% of the release time data. BPINNs provide tighter and more reliable uncertainty quantification under noise. By combining physical laws with experimental data, the proposed framework yields highly accurate long-term release predictions from short-term measurements, offering a practical route for accelerated characterization and more efficient early-stage drug release system formulation.
\end{abstract}

\end{frontmatter}

\section*{Introduction}

Controlled release plays a pivotal role in modern medicine and pharmacology, offering the potential to target therapeutic agents with precision, thereby optimizing efficacy and minimizing side effects. The controlled release of drugs is essential in numerous clinical contexts, from cancer therapy \cite{zahra2023} to the management of chronic diseases \cite{langer1990}. Effective drug delivery systems ensure that drugs reach their intended targets at the correct dosage and over the appropriate time frame, a critical factor in improving patient outcomes. The design and simulation of such systems rely heavily on modeling drug transport mechanisms within various substrates, including polymers and tissues \cite{brannonpeppas1995}.

The mathematical modeling of drug release is grounded in several classical models, with Fick's Law of Diffusion, Higuchi's Model, and the Peppas-Korsmeyer Model being some of the foundational approaches. Fick's Law describes the diffusion process based on concentration gradients and is widely used to model drug transport through porous matrices \cite{crank1979mathematics}. The Higuchi model, initially developed for planar systems, expands upon Fickian diffusion to describe the release of drugs from thin films, taking into account the square-root time dependency \cite{higuchi1961rate}. Peppas's Model, on the other hand, generalizes drug release behavior, including both Fickian and non-Fickian diffusion processes, allowing for more versatile characterization of drug release kinetics in various geometries and materials \cite{ritger1987simple}. These models have served as the backbone of drug release modeling for decades, providing analytical solutions for specific release system geometries.

However, real-world drug release systems are often more complex, necessitating the use of modeling assumptions and numerical methods to solve partial differential equations (PDEs) where analytical solutions are not feasible \cite{siepmann2008mathematical}. Traditional numerical methods, such as finite element and finite difference methods, have been applied to solve these models under different boundary and initial conditions \cite{xu2013mathematical}. The limitations of these classical approaches become more clear in complex material systems, such as hydrogels, where factors such as polymer concentration and structural interactions influence drug release \cite{shukla2020effect}. Recently, neural networks have demonstrated success in addressing broader mechanistic problems \cite{niu2022ultrasound,Niu2022UltrasoundNetwork}, as well as in applications related to targeted drug delivery \cite{abdalla2024machine}\cite{castro2021machine}.  In particular, Physics-Informed Neural Networks (PINNs), have emerged as powerful tools for solving initial and boundary value mechanistic problems \cite{raissi2019physics} \cite{niu2023modeling,konale2025PINN}. PINNs leverage deep learning to integrate physical laws, described by differential equations, into the training process, allowing neural networks to approximate solutions to complex PDEs \cite{karniadakis2021physics}. In addition to PINNs, Bayesian Physics-Informed Neural Networks (BPINNs) provide a framework for handling noisy data by incorporating uncertainty quantification into the predictions, which is crucial when dealing with real-world, imperfect data \cite{yang2021b}. This allows BPINNs to simultaneously solve forward and inverse problems with greater robustness. Our main contribution is to propose a method beyond classical drug release models that integrates a small amount of real-life/experimental data with physical laws to inform neural networks that can then make accurate drug release predictions. This hybrid approach (i) achieves higher predictive accuracy than traditional methods and (ii) requires significantly fewer data points, enabling reliable drug release modeling even under limited experimental conditions.

To address the inverse problem in drug delivery, which is often ill-posed and challenging to solve using conventional methods, we leverage PINNs. Inverse problems typically involve determining unknown parameters or functions from observed data. These inverse problems have been applied to mechanics \cite{shukla2021physics}, but in drug delivery, this data can be incomplete, noisy, or poorly defined, making traditional solutions unstable or unreliable. PINNs offer a solution by enabling a "one-shot" approach, where the neural network learns both the solution to the differential equation and the parameters of interest simultaneously. However, in scenarios where experimental data has significant noise, the deterministic nature of PINNs can lead to large variances and less accurate results. This is where BPINNs can help by incorporating Bayesian inference to manage uncertainty and improve model robustness in the presence of noise. Unlike traditional numerical solvers, which struggle with noisy data, BPINNs enable uncertainty quantification, providing a probabilistic framework that can yield reliable solutions even with imperfect data. The Bayesian approach, though computationally intensive due to methods like Hamiltonian Monte Carlo (HMC) sampling, offers significant advantages in handling noisy, real-world experimental data, especially in higher-dimensional problems \cite{yang2021b}.

This study leverages the experimental data by Liu et al. \cite{liu2021}, which explored the controlled release of molecular intercalants from two-dimensional nanosheet films. In their study, different types of films (flat, 1D, and 2D) were used to study drug release, providing critical data for modeling diverse drug release systems. Utilizing the data from this study, we introduce a new machine learning based approach that integrates experimental data with approximate physical laws to yield more accurate predictions of drug release.

The scope of this paper is to develop and compare both deterministic and probabilistic approaches to modeling drug release using PINNs and BPINNs. By training an ensemble of PINNs and utilizing BPINNs with uncertainty quantification, we introduce a new methodology that enhances the predictive accuracy of drug release models across various film types—flat, 1D wrinkled, and 2D crumpled. Our approach of using classical models, deep learning, and uncertainty quantification provides a more robust framework for the simulation of drug delivery systems. Moreover, this study highlights the capability of PINNs, using Fick's law as the PDE, to achieve high predictive accuracy with minimal data points, making them especially valuable in early-stage drug development and scenarios where experimental data is limited. The results of this work provide motivation for extending and applying PINN-based approaches in controlled drug release and to the broader biotechnology applications.

\section*{Methods}

\subsection*{Data Acquisition and Preprocessing}
\begin{figure}[h!]
    \centering
    \includegraphics[width=0.8\textwidth]{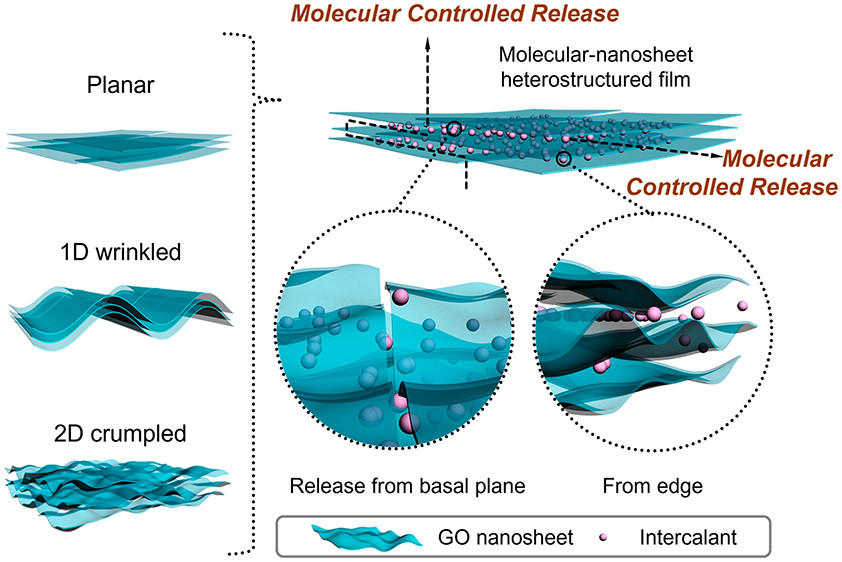}
    \caption{Reprinted (adapted) with permission from \cite{liu2021}. Copyright 2021 American Chemical Society. The figure shows the molecular release mechanisms in different nanosheet films. 
        The figure displays three types of films used in the controlled molecular release study: 
        \textbf{Planar}, \textbf{1D wrinkled}, and \textbf{2D crumpled}. 
        Each structure facilitates a unique release profile:
        \textbf{Planar films} allow a uniform release from the basal plane, while 
        \textbf{1D wrinkled} and \textbf{2D crumpled films} enable more complex, controlled release, especially from the film edges, creating a heterogeneous release profile.
        The graphene oxide nanosheet structures serve as carriers for intercalant molecules, with release observed from both the basal plane and edges.
    }
    \label{fig:liu_architecture}
\end{figure}

The drug release data employed in this study originates from the work by Liu et al. \cite{liu2021}, which investigated molecular release profiles from two-dimensional graphene oxide (GO) nanosheet films with intercalated agents. Liu et al. developed three distinct film architectures—planar, 1D wrinkled, and 2D crumpled structures—to study how film morphology affects the release behavior of a model compound, Rhodamine B, into solution. These configurations, visualized in Figure~\ref{fig:liu_architecture}, demonstrate increasing complexity in structural design, intended to modulate release dynamics by altering diffusion pathways.

In the experimental setup, GO films were cast onto polystyrene substrates and thermally treated to achieve the desired wrinkled or crumpled patterns. The films were then immersed in a buffered solution, and cumulative release was monitored by measuring Rhodamine B concentrations over time through UV-visible spectrophotometry. The concentration values were recorded at regular intervals with sufficient precision to capture both the rapid burst release at early times and the slower sustained release at later times. This level of temporal resolution makes the dataset well-suited for evaluating model accuracy across different release phases.

However, as with any experimental data, the measurements may contain inherent error due to environmental variability, instrument precision limits, and operator handling. To reflect such real-world measurement scenarios, we introduced Gaussian noise into the dataset, simulating potential experimental errors. This augmentation improves model robustness by forcing it to learn under conditions closer to practical laboratory setups. The quality of the experimental data is reflected in its granularity and consistent sampling intervals, which capture both early burst release and later sustained release dynamics. The noise simulation procedure and its role in model training are described in detail in the noise simulation section.

\subsection*{Classical Drug Time Release Models}
Three classical models were used to benchmark time-history release of drugs from films: Fick's Law, Higuchi’s model, and Peppas’s model.

\subsubsection*{Fick's Second Law of Diffusion}
Fick's Second Law governs the time-dependent diffusion of drugs and is 
expressed as:

\begin{equation}
\frac{\partial u}{\partial t} = D \nabla^2 u,
\label{eq:ficks_law}
\end{equation}

where \( u \) represents the drug concentration at a point in space, \( t \) is time, \( D \) is the diffusion coefficient, and \( \nabla^2 \) is the Laplace operator, accounting for spatial diffusion in the system. This law models the diffusion-driven release of drugs, particularly useful for homogeneous, non-reactive systems where drug transport is dominated by diffusion across the film. Fick's Second Law forms the backbone of many classical diffusion-based {approximate} models in drug release studies and serves as a baseline for comparison against more complex models, such as Higuchi's and Peppas's mentioned after \cite{crank1979mathematics}.

\subsubsection*{Higuchi’s Model}
Higuchi’s model accounts for drug release from a semi-infinite matrix thin film (patch) and is based on a diffusion-controlled process. Higuchi \cite{higuchi1961rate} described it as:

\begin{equation}
Q = A \cdot \sqrt{D \cdot (2C_s - C)} \cdot t^{1/2},
\label{eq:higuchi}
\end{equation}

where \( Q \) is the amount of drug released, \( A \) is a constant, \( D \) is the diffusion coefficient, \( C_s \) is the solubility of the drug in the matrix, and \( C \) is the initial drug concentration \cite{higuchi1961rate}. This model assumes that the concentration gradient drives the release process and is valid for the early stages of diffusion.

\subsubsection*{Peppas’s Model}
Peppas’s model generalizes the drug release process, particularly accounting for both Fickian diffusion and non-Fickian mechanisms through the empirical power-law equation:

\begin{equation}
\frac{M_t}{M_\infty} = k \cdot t^n
\label{eq:peppas}
\end{equation}

where \( M_t/M_\infty \) is the fraction of drug released at time \( t \), \( k \) is a kinetic constant, and \( n \) is the diffusion exponent that characterizes the release mechanism. This model extends Fickian diffusion to include more complex release behaviors such as swelling or degradation of the delivery matrix \cite{ritger1987simple}.

\subsection*{Physics-Informed Neural Networks (PINNs)}
PINNs were used to solve the diffusion equations while incorporating the underlying physics of Fick’s Law, together with the experimental data. The network takes normalized time and spatial coordinates as inputs and outputs predicted drug concentrations. The architecture consists of fully connected layers, with a hybrid loss function that enforces agreement with both Fick’s Law and the available experimental measurements \cite{raissi2019physics}. The loss function was defined as,
\[
\mathcal{L} = \mathcal{L}_{data} + \mathcal{L}_{physics},
\]
where \( \mathcal{L}_{data} \) minimizes the error between the predicted and actual concentrations, and \( \mathcal{L}_{physics} \) penalizes deviations from Fick’s Law. Note that the function is expressed as:
\[
    f(x, t) = \frac{\partial u}{\partial t} - D \frac{\partial^2 u}{\partial x^2}.
\]

As seen in Figure \ref{fig:pinn_bnn_architectures}a, the PINN is a hybrid model due to the use of both Fick's law and data behavior. Naturally, this is better than Fick's law since it incorporates both principles and supplements Fick's law when it is not accurate.

\subsection*{Bayesian Physics-Informed Neural Networks (BPINNs)}

\begin{figure*}[h!]
    \centering
      \includegraphics[width=\textwidth,height=0.6\textheight,
            keepaspectratio]{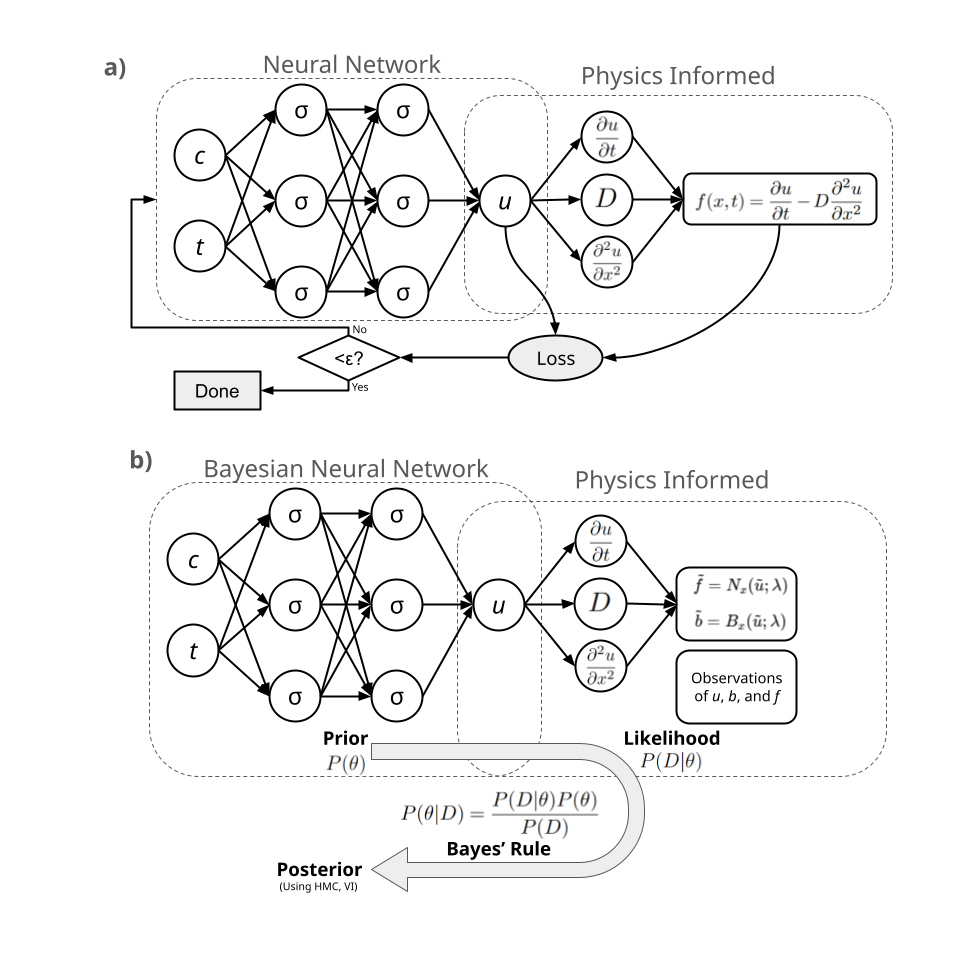}
    \caption{
        Detailed PINN and BPINN architectures. 
        (a) The standard PINN framework integrates a neural network with physical laws. Inputs, represented by concentration \( c \) and time \( t \), are fed into a fully-connected neural network with activation functions (\( \sigma \)) in each hidden layer. The network predicts \( u \), which is subjected to physical constraints based on {data driven loss and} Fick’s law of diffusion: \( f(x, t) = \frac{\partial u}{\partial t} - D \frac{\partial^2 u}{\partial x^2} \), where \( D \) denotes the diffusion coefficient. This physical loss is computed by applying automatic differentiation to enforce the equation at collocation points. The output \( f(x, t) \) is then used to calculate the total loss, combining both data-driven loss and physics-based loss. The training loop iterates until the total loss falls below a predefined threshold \( \epsilon \), at which point the model is considered trained and outputs a “Done” signal.
        (b) The BPINN extends the PINN by incorporating Bayesian inference, allowing for uncertainty quantification in the model parameters. In this architecture, a Bayesian Neural Network (BNN) with inputs \( c \) and \( t \) generates probabilistic predictions of \( u \), modeling each weight and bias as a distribution rather than a fixed value. The physical law, represented as the likelihood \( P(D|\theta) \), is enforced by the physics-informed part of the network. Here, the system dynamics are governed by equations for \( \tilde{f} = N_x(\tilde{u}; \lambda) \) and \( \tilde{b} = B_x(\tilde{u}; \lambda) \), where \( \tilde{u} \) represents predictions with uncertainty and \( \lambda \) are additional parameters representing model priors. Observations \( u \), \( b \), and \( f \) provide data-driven grounding. The Bayesian inference is applied through Bayes' rule: \( P(\theta|D) = \frac{P(D|\theta)P(\theta)}{P(D)} \), where \( P(\theta) \) is the prior derived from the BNN and \( P(D|\theta) \) is the physics-informed likelihood. The posterior \( P(\theta|D) \) is obtained via sampling techniques such as Hamiltonian Monte Carlo (HMC) or Variational Inference (VI), providing a distribution over the model parameters.}
    \label{fig:pinn_bnn_architectures}
\end{figure*}

Bayesian Physics-Informed Neural Networks (B-PINNs) integrate the traditional PINN framework with Bayesian Neural Networks (BNNs) \cite{bykov2021explaining} to enable uncertainty quantification in predictions \cite{yang2021b}. This hybrid framework combines the strengths of BNNs \cite{bishop1997bayesian} and PINNs to address both forward and inverse problems involving nonlinear dynamics. As illustrated in Figure~\ref{fig:pinn_bnn_architectures}b, prior distributions are placed over the neural network parameters, while the physics-informed component enforces Fick’s law through a physical loss term and incorporates experimental observations through a data loss term. Bayes’ theorem is then applied to combine the prior and likelihood, yielding posterior distributions over both the ODE parameters and network weights. This setup provides a robust methodology for handling problems with uncorrelated noise and enables both aleatoric and epistemic uncertainty quantification in the learned parameters. 

The BNN component incorporates Bayesian principles by placing probability distributions over the neural network's weights and biases. To account for data noise, we include a noise term in the likelihood function. Applying Bayes’ rule then allows us to estimate the posterior distributions of both the model and the ODE parameters. This process facilitates the propagation of uncertainty from the observed data through to the model's predictions.
We express Equation~\eqref{ODE} as:
\begin{subequations}
\label{ODE}
\begin{align}
& \mathcal{N}_t(r; \bm{\lambda}) = f(t),\quad t \in \mathbb{R}^+ \\  
& \mathcal{I} (r, \bm{\lambda}) = r_0, \quad t=0,
\end{align}    
\end{subequations}
Here, $\bm{\lambda}$ denotes the vector of parameters in the ODE (Equation~\eqref{ODE}), which in our case includes $D$. The operator $\mathcal{N}_t$ represents a general differential operator, $f(t)$ is the forcing term, and $\mathcal{I}$ specifies the initial condition. This setup defines an inverse problem, where the parameters $\bm{\lambda}$ are inferred from observed data, along with estimates of both aleatoric and epistemic uncertainties.
The likelihoods associated with the simulation data and ODE parameters are defined as:
\small
\begin{align}\label{liklihood}
 \nonumber
\!\! P(\mathcal{D} \! \mid \! \boldsymbol{\theta}, \bm{\lambda}) & \! = \! P\left(\mathcal{D}_{r} \mid \boldsymbol{\theta}\right) P\left(\mathcal{D}_f \mid \boldsymbol{\theta}, \bm{\lambda}\right) P\left(\mathcal{D}_{\mathcal{I}} \mid \boldsymbol{\theta}, \bm{\lambda}\right),~\text{where} \\  \nonumber
\! \! \! \! \! P\! \left(\mathcal{D}_{r} \! \mid \! \boldsymbol{\theta}, \bm{\lambda}\right) & \! = \!\prod_{i=1}^{N_{r}} \!  \frac{1}{\sqrt{2 \pi \sigma_r^{(i)^2}}} \exp \! \! \left[\! -\frac{\left({r}(\boldsymbol{t}_r^{(i)} ; \boldsymbol{\theta}, \boldsymbol{\lambda})-\bar{r}^{(i)}\right)^2}{2 \sigma_r^{(i)^2}}\right]\!\!, \\  \nonumber
\! \! \! \! \! P\! \left(\mathcal{D}_f \! \mid \! \boldsymbol{\theta}, \bm{\lambda}\right) & \! = \!\prod_{i=1}^{N_f} \!  \frac{1}{\sqrt{2 \pi \sigma_f^{(i)^2}}} \exp \! \! \left[\! -\frac{\left(f(\boldsymbol{t}_f^{(i)} ; \boldsymbol{\theta}, \boldsymbol{\lambda})-\bar{f}^{(i)}\right)^2}{2 \sigma_f^{(i)^2}}\right]\!\!, \\  
\!\! \! \! \! \! P\!\left(\mathcal{D}_\mathcal{I} \! \mid \! \boldsymbol{\theta}, \boldsymbol{\lambda}\right) & \! = \!\prod_{i=1}^{N_\mathcal{I}} \! \frac{1}{\sqrt{2 \pi \sigma_\mathcal{I}^{(i)^2}}} \exp \! \! \left[ \!-\frac{\left(\mathcal{I}(\boldsymbol{t}_i^{(i)} ; \boldsymbol{\theta}, \boldsymbol{\lambda}) \! - \! \bar{\mathcal{I}}^{(i)}\right)^2}{2 \sigma_{\mathcal{I}}^{(i)^2}}\right]\!\!,
\end{align}
\normalsize
where $D\! = \!D_r \cup D_f \cup D_{\mathcal{I}}$ with 
$\mathcal{D}_r\! = \!\left\{\left({\ln t}_r^{(i)}, \bar{\ln r}^{(i)}\right)\right\}_{i=1}^{N_r}$, 
$\mathcal{D}_f\! = \!\left\{\left(t_f^{(i)}, f^{(i)}\right)\right\}_{i=1}^{N_f}$, 
$\mathcal{D}_\mathcal{I}\! = \!\left\{\left(t_\mathcal{I}^{(i)}, \mathcal{}{I}^{(i)}\right)\right\}_{i=1}^{N_\mathcal{I}}$ are scattered noisy measurements.
The joint posterior of $[\bm{\theta}, \bm{\lambda}]$ is given as
\begin{align}\label{posterior}
\begin{aligned}
P(\boldsymbol{\theta}, \boldsymbol{\lambda} \mid \mathcal{D})&=\frac{P(\mathcal{D} \mid \boldsymbol{\theta}, \boldsymbol{\lambda}) P(\boldsymbol{\theta}, \boldsymbol{\lambda})}{P(\mathcal{D})}\\
&\simeq P(\mathcal{D} \mid \boldsymbol{\theta}, \boldsymbol{\lambda}) P(\boldsymbol{\theta}, \boldsymbol{\lambda})\\
&=P(\mathcal{D} \mid \boldsymbol{\theta}, \boldsymbol{\lambda}) P(\boldsymbol{\theta}) P(\boldsymbol{\lambda}).
\end{aligned}
\end{align}

To sample parameters from the posterior probability distribution defined by Equation~\eqref{posterior}, we utilized the Hamiltonian Monte Carlo (HMC) approach \cite{radivojevic2020modified}, which is an efficient Markov Chain Monte Carlo (MCMC) method \cite{brooks1998markov}. For a detailed explanation of this method, please refer to, e.g., Refs. \cite{neal2011mcmc, neal2012bayesian, graves2011practical}. Alternatively, variational inference \cite{blei2017variational} can be employed to approximate the posterior distribution. In variational inference, the posterior density of the unknown parameter vector is approximated by another parameterized density function restricted to a smaller family of distributions \cite{yang2021b}. To compute uncertainty in the ODE parameters using B-PINN, we added 5\% noise to the original dataset, where the noise was sampled from a normal distribution with mean zero and a standard deviation of ±1.

B-PINNs, which extend the traditional PINN framework, were used to quantify uncertainty in model predictions. By integrating Bayesian inference via Monte Carlo Dropout, B-PINNs estimate predictive uncertainty by sampling multiple dropout configurations during forward passes \cite{yang2021b}. Each B-PINN model was trained under the same conditions as the PINN models, with dropout applied after every hidden layer.

For each film type, 100 forward passes were performed to calculate the mean and standard deviation of the predicted drug concentrations, enabling uncertainty quantification through:
\[
\hat{u} = \frac{1}{N} \sum_{i=1}^N u_i
\]
\[
\sigma_u^2 = \frac{1}{N} \sum_{i=1}^N (u_i - \hat{u})^2
\]
where \( u_i \) is the prediction from the \( i \)-th forward pass, and \( N \) is the total number of passes. This allowed the calculation of both the mean prediction and uncertainty bounds.

\subsection*{Training and Evaluation Metrics}
The models were trained using {10,000 collocation points, chosen empirically to balance accuracy and computational cost. Tests with fewer points (1,000–5,000) led to under-constrained solutions, while larger values gave little improvement but increased training time.} Collocation points were {sampled with Latin Hypercube Sampling (LHS) for broad input coverage,} enforcing adherence to Fick’s Law as part of the physics-informed loss, while a data-driven loss ensured model predictions closely fit the experimental data from Liu et al. \cite{liu2021}.

In the case of PINNs, the training process was conducted across three film types (Flat, 1D wrinkled, and 2D crumpled) with a standard architecture of five hidden layers and 20 neurons per layer, using a learning rate of $1 \times 10^{-3}$. For BPINNs, training was repeated with dropout layers added to estimate uncertainty, and results were aggregated over 50 runs to simulate ensemble behavior.

The training setups varied as follows:
\begin{itemize}
    \item \textbf{PINN vs. Classical Models}: To compare against traditional methods, PINNs were trained for each film type, focusing on minimizing both physics-informed and data-driven losses.
    \item \textbf{BPINN vs. Ensemble of 50 PINNs}: The BPINN models utilized a Bayesian framework with dropout layers to assess uncertainty. Comparisons with an ensemble of 50 PINNs highlighted BPINN’s robustness in uncertainty quantification.
    \item \textbf{Training PINNs on Limited Data}: The models were also trained with progressively limited data points to assess predictive and generalization capabilities in data-scarce scenarios.
\end{itemize}

The performance of both PINNs and BPINNs was evaluated using the following metrics:
\begin{itemize}
    \item \textbf{Mean Absolute Error (MAE)}: This metric captures the average magnitude of prediction errors, providing an intuitive measure of accuracy across film types. Mathematically, MAE is defined as:
    \begin{equation}
        \text{MAE} = \frac{1}{n} \sum_{i=1}^{n} \left| y_i - \hat{y}_i \right|,
    \end{equation}
    where \( y_i \) is the true value, \( \hat{y}_i \) is the predicted value, and \( n \) is the total number of data points.

    \item \textbf{Root Mean Square Error (RMSE)}: By emphasizing larger errors, RMSE offers insight into the model’s sensitivity to outliers, especially beneficial for assessing fit in regions with sparse or high-variance data. Mathematically, RMSE is defined as:
    \begin{equation}
        \text{RMSE} = \sqrt{\frac{1}{n} \sum_{i=1}^{n} \left( y_i - \hat{y}_i \right)^2}.
    \end{equation}
\end{itemize}

A summary of the architecture parameters and hyperparameters used in each training setup is provided in Table \ref{tab:pinn_bipinn_training}.

\begin{table*}[h!] 
    \centering
    \resizebox{\textwidth}{!}{ 
    \begin{tabular}{|l|c|c|c|c|}
        \hline
        \textbf{Parameter/Setup} & \textbf{Classical PINN Training} & \textbf{50 PINNs Training} & \textbf{BPINN Training} & \textbf{Limited Data Training} \\
        \hline
        Number of Layers & 5 & 5 & 5 & 5 \\
        Neurons per Layer & 20 & 20 & 20 & 20 \\
        Input Dimension & 2 (time, concentration) & 2 (time, concentration) & 2 (time, concentration) & 2 (time, concentration) \\
        Output Dimension & 1 (solution \(u\)) & 1 (solution \(u\)) & 1 (solution \(u\)) & 1 (solution \(u\)) \\
        Activation Function & tanh & tanh & tanh (with dropout) & tanh \\
        Initializer & hyper\_initial & hyper\_initial & hyper\_initial & hyper\_initial \\
        Optimizer & Adam & Adam & Adam & Adam \\
        Learning Rate & 1e-3 & 1e-3 & 1e-3 & 1e-3 \\
        Epochs & 2500 & 5000 & 10000 & 2000 \\
        Collocation Points (\(N_f\)) & 10000 & 10000 & 10000 & 1000 \\
        Diffusion Coefficient (\(D\)) & 0.01 & 0.01 & 0.01 & 0.01 \\
        Uncertainty Estimation & None & Mean \& Std over 50 runs & Dropout-based (Monte Carlo) & None \\
        Data Points Used & All & All & All & Incremental from 2 to 14 \\
        \hline
    \end{tabular}
    }
    \caption{Summary of architecture parameters and hyperparameters for different training setups in the PINN and BPINN experiments.}
    \label{tab:pinn_bipinn_training}
\end{table*}

\subsection*{Improving on Classical Models}
The primary goal of this study is to propose a new method for predicting drug delivery that performs better than classical models (Fick's Law, Higuchi’s, and Peppas’s) by integrating data-driven methods through PINNs and BPINNs. Classical models assume idealized conditions and cannot account for experimental noise or complex boundary conditions. By embedding the physical laws into neural networks, we achieve a hybrid approach that retains the interpretability of classical methods while enhancing accuracy through data fitting and uncertainty quantification. 

We first fit all of the classical models to the experimental drug release data across the three film types (Flat, 1D, and 2D). For each model, parameters such as the diffusion coefficient \( D \), and for Peppas’s model, the kinetic constant \( k \) and diffusion exponent \( n \), were estimated using curve fitting techniques. The error between the classical model predictions and the experimental data was quantified using Mean Absolute Error (MAE) and Root Mean Square Error (RMSE). Once these baselines were established, we trained Physics-Informed Neural Networks (PINNs) on the same datasets to model the drug release profiles. Each PINN was constrained by Fick's law and data in the loss function, and was trained to predict drug concentration over time. This improved upon the classical models by incorporating the physics and data both directly into the learning process. To evaluate comparative predictive accuracy and goodness-of-fit, we calculated the MAE and RMSE and plotted the values against each other.

\subsection*{Performance of PINN and BPINN for noisy data} 
Real-world data often contains inherent noise due to experimental variations, which classical models fail to address effectively. To simulate such real-world conditions, Gaussian noise was introduced into the training data for each of the film types, allowing us to model the variability observed in actual experimental conditions. The noisy data enabled the PINNs to generalize better to unseen conditions, leading to more robust predictions.

For each of the film types, 50 PINNs were trained, each using a different randomly initialized set of weights and a unique noisy dataset generated through Gaussian noise (mean \( \mu = 0 \), standard deviation \( \sigma = 0.1 \)). These models were trained for 5000 epochs using a combination of data-driven losses and physics-based losses. The choice of 50 PINNs and 5000 training epochs ensures a robust ensemble that can capture variability in predictions due to random initialization and noise, while providing sufficient training time for convergence. The Gaussian noise parameters (\( \mu = 0 \), \( \sigma = 0.1 \)) are selected to realistically simulate experimental uncertainty without overwhelming the model’s learning capacity.

After training the PINN ensemble, a BPINN was trained for each film type, using Monte Carlo Dropout to quantify predictive uncertainty \cite{yang2021b}. The BPINNs were similarly trained with noisy datasets, but they provide additional information by predicting the standard deviation of the output in addition to the mean, allowing for uncertainty quantification.
Finally, the predictions from the PINN ensemble and BPINN were compared to the experimental data, and their performance was evaluated using MAE and RMSE. These metrics allowed us to assess the improvements of the hybrid models over classical methods, as well as the benefits of incorporating Bayesian inference to handle noisy data.

\subsection*{PINN with Limited Data Points}

To assess the efficacy of predictive models on drug delivery profiles, we implemented a methodology focused on the use of minimal data points to simulate the real-world constraint of limited experimental data. The PINN consisted of Fick's law as the physics-informed loss. The models were trained incrementally on increasing numbers of data points \( n \), from 2 to 14, with the goal of determining the minimum required data to achieve a robust prediction. The total number of data points was 15. 

We calculated the RMSE as our primary performance metric, which emphasizes larger deviations and thus provides an indication of model reliability in a sensitive drug release setting.  Each model was trained on noisy data for each film type, and RMSE values were computed to identify the best-performing model at each data point increment.

\section*{Results and Discussion}

In this section, we discuss the performance of drug release models from classical models, PINNs, and BPINNs. We begin by evaluating the classical models across different film types, highlighting the limitations these models face in capturing the complexities of drug release in more structured film configurations, such as 1D and 2D films. Next, we compare PINNs with classical models, highlighting the advantages of integrating physical laws with machine learning to achieve more accurate and flexible predictions. Further, we examine the robustness of PINNs and BPINNs under noisy data conditions, showcasing the BPINN’s enhanced handling of uncertainty and variability. Lastly, we explore model performance in data-limited scenarios, demonstrating the adaptability and efficiency of PINNs in providing reliable predictions with minimal data, which is crucial for early-stage drug development applications. Each subsection provides a detailed analysis of these findings, supported by relevant figures and error metrics (MAE and RMSE) to substantiate our results.

\subsection*{Performance of Classical Models}
The performance of the classical drug release models—Higuchi's, Peppas's, and Fick's Law—varied notably across the different film types. Higuchi's and Peppas's models demonstrated relatively better fits for the Flat Film, particularly due to their formulation’s capacity to account for diffusion mechanisms in systems with constant diffusivity and non-swelling matrices. However, these models exhibited clear limitations when applied to the 1D wrinkled and 2D crumpled films, as the drug release profiles became more complex due to the higher dimensionality and varying boundary conditions.

Fick’s Law, on the other hand, showed superior predictive capabilities for the 1D wrinkled and 2D crumpled films. This is expected, as Fick’s Law is rooted in the fundamental principle of diffusion, making it more appropriate for cases where drug release is governed primarily by concentration gradients. Yet, even with this theoretical advantage, Fick’s Law struggled to account for non-ideal behaviors observed in the experimental data, such as variability in diffusivity or heterogeneous film structures, which are common in real-world applications \cite{crank1979mathematics}. 

\begin{figure}[h!]
    \centering
    \includegraphics[width=0.8\textwidth]{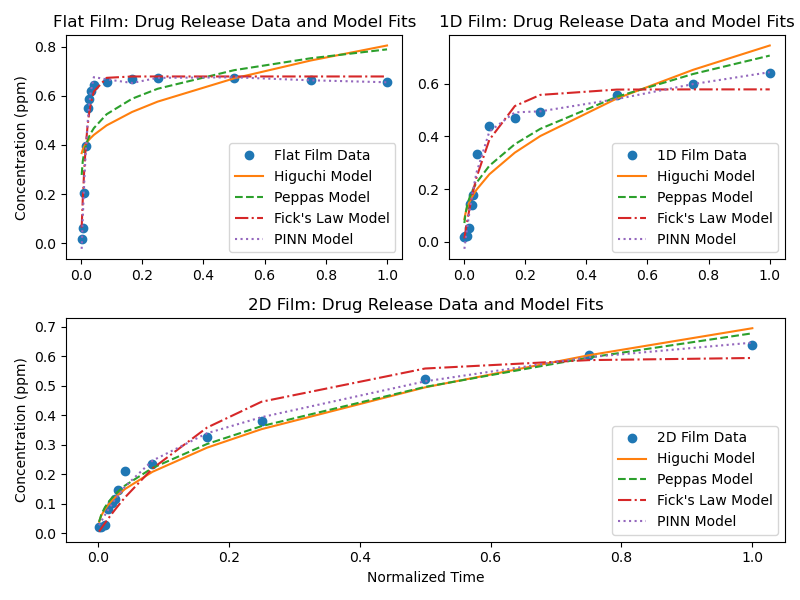}
    \caption{Drug release data for flat, 1D, and 2D films fitted using classical models and PINNs. (Top left) The flat film data shows the release profile of drug concentration over normalized time. Each model attempts to capture the diffusion characteristics and release of therapeutics, with the PINN model closely aligning with experimental data points. (Top right) The 1D wrinkled film data, which adds a layer of structural complexity, shows a different release profile. Classical models fit the data, but the PINN model’s flexibility allows it to accommodate the nuanced release dynamics more effectively. (Bottom) The 2D crumpled film data represents the most complex structure, with multi-dimensional diffusion effects. The PINN model maintains a consistent fit across the release profile, capturing both rapid initial release and slower sustained release phases, while classical models demonstrate limitations. The total time (normalized time = 1) for all release curves is 48 hours. }
    \label{fig:plot_pinn}
\end{figure}

\begin{figure}[h!]
    \centering
    \includegraphics[width=0.75\textwidth]{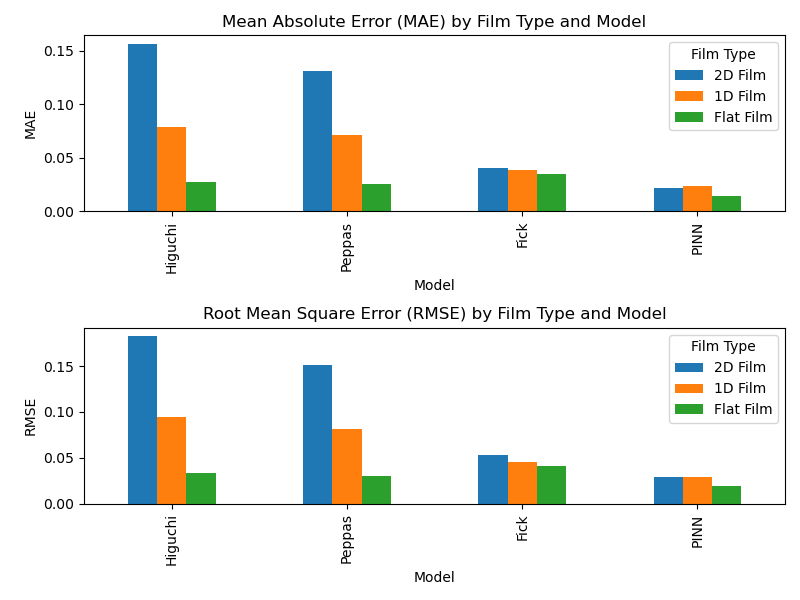}
    \caption{MAE and RMSE for classical models and PINNs across all film types (Flat, 1D, 2D). (Top) MAE values for each model indicate the average prediction errors for drug release across the three film types. Higher MAE values in the Higuchi and Peppas models for the 2D film reveal their limitations in capturing complex release patterns, while the PINN achieves low errors across all films, indicating consistent accuracy. (Bottom) RMSE values provide insight into error sensitivity to outliers. Similar to the MAE, RMSE is higher for classical models in the 2D film, highlighting challenges with structural complexity. The PINN model demonstrates robustness with low RMSE across all films, reflecting its resilience against large deviations.}
    \label{fig:error_pinn}
\end{figure}

However, the integration of both Fick’s Law and the experimental data into the PINN loss function yielded consistently better results across all film types. As demonstrated in Figure~\ref{fig:plot_pinn}, this hybrid loss formulation allowed the PINN to outperform all classical models, providing a more accurate fit to the experimental release profiles for the Flat, 1D, and 2D films. The total time (normalized time = 1) for all release curves is 48 hours. The error analysis in Figure~\ref{fig:error_pinn} further reinforces this conclusion, with the PINN achieving significantly lower MAE and RMSE across all cases. 

This demonstrates that combining physics-based laws with machine learning allows PINNs to capture complexities in drug release that classical models often miss. Classical methods, while effective in ideal conditions, are limited by assumptions like uniform diffusion and simple geometries, which are rarely met in practice. In contrast, PINNs adapt to the system's intricacies through data-driven learning while maintaining the underlying physical laws. The PINN's superior performance across all film types underscores the strength of this hybrid approach, offering more accurate and reliable predictions in drug delivery contexts.

\subsection*{Simulating Noise}

Real-world and experimental data often contain noise from various sources, such as measurement errors and fluctuating environmental conditions. To simulate these real-world variations and assess model robustness, we introduced Gaussian noise into the drug release datasets for all three film types. The added noise simulates common challenges in biological systems, emphasizing the need for robust models that can generalize effectively to noisy, unseen data.

\begin{figure}[h!]
    \centering
    \includegraphics[width=0.8\textwidth]{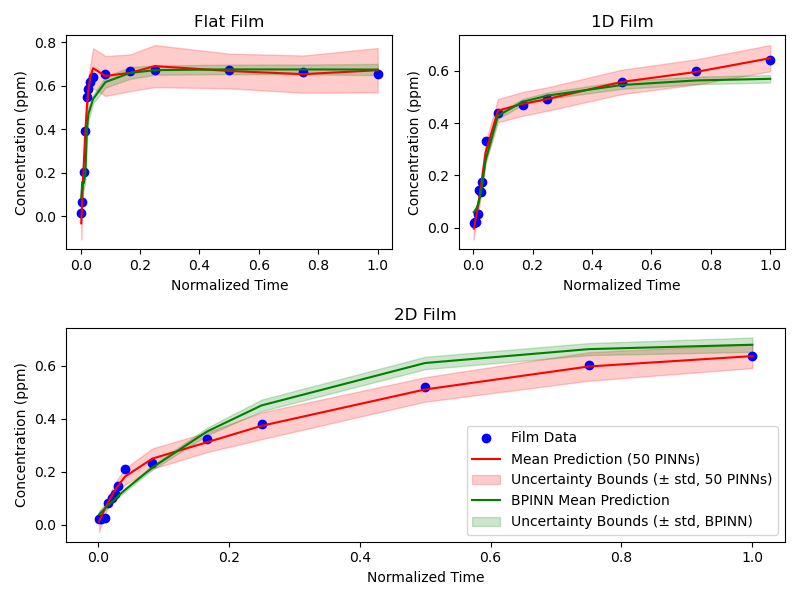}
    \caption{PINN and BPINN predictions with uncertainty bounds for simulated noisy data across flat, 1D, and 2D films. 
    (Top left) Flat film: {Both PINN (red) and BPINN (green) capture the overall release profile, with BPINN producing narrower uncertainty bands.} 
    (Top right) 1D film: PINN and BPINN predictions remain close to the data, though BPINN again shows reduced variance.
    (Bottom) 2D film: The PINN ensemble aligns more closely with the experimental data on average, while the BPINN provides tighter uncertainty bounds but shows a small bias. Overall, BPINNs improve robustness to noise by quantifying uncertainty, whereas PINNs may better capture mean behavior in some cases. The total time (normalized time = 1) for all release curves is 48 hours. }
    \label{fig:plot_noise}
\end{figure}

To evaluate the model under noisy conditions, each of the 50 PINNs was trained with different initializations, creating an ensemble of predictions for each film type. Figure~\ref{fig:plot_noise} illustrates these predictions, where the red-shaded uncertainty bounds represent the ensemble variance. The BPINN predictions, shown in green, demonstrate narrower uncertainty bounds, effectively capturing the noisy data with greater precision.

While the standard PINNs handle noiseless data well, they exhibit higher sensitivity to noise, resulting in significant variance in parameter estimations, especially as noise levels increase. This sensitivity leads to wider uncertainty bands in the predictions, limiting the reliability of PINNs in high-noise scenarios. BPINNs, by contrast, are specifically designed to manage data noise through Bayesian inference, incorporating Hamiltonian Monte Carlo (HMC) sampling to estimate model parameters and quantify uncertainty effectively. This approach yields more reliable predictions and reduced variance in parameter estimates, even under substantial noise.

However, the application of BPINNs is computationally expensive due to the inherent complexity of HMC sampling, which requires substantial computational resources and time. This limitation becomes increasingly pronounced in higher-dimensional data, where the cost of sampling scales significantly, potentially impacting BPINN scalability.

\begin{figure}[h!]
    \centering
    \includegraphics[width=0.8\textwidth]{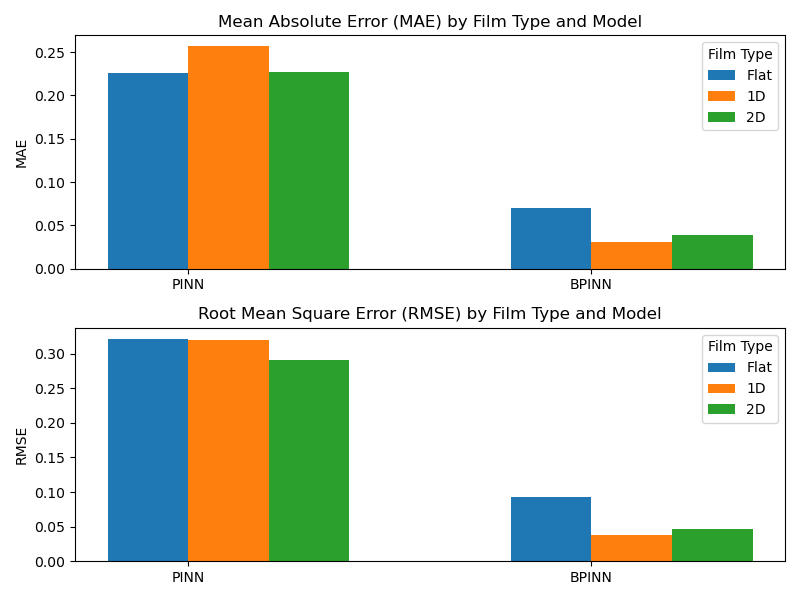}
    \caption{MAE and RMSE for PINNs and BPINNs across film types (Flat, 1D, 2D) under noisy conditions. (Top) MAE comparison: BPINNs consistently demonstrate lower MAE values across all film types, indicating greater accuracy in capturing the drug release profile despite the presence of noise. PINNs, on the other hand, exhibit higher MAE, particularly for the 1D and 2D films, reflecting their sensitivity to noise. (Bottom) RMSE comparison: BPINNs achieve significantly reduced RMSE values, highlighting their robustness against outliers and greater stability across film types. This reduction in RMSE suggests BPINNs’ enhanced ability to generalize to noisy data scenarios compared to PINNs, where RMSE remains consistently higher across all film types.}
    \label{fig:error_noise}
\end{figure}

Figure~\ref{fig:error_noise} quantifies model performance through MAE and RMSE metrics, highlighting the BPINN's superior robustness over PINNs, particularly in noisy settings. The BPINN's capacity to account for uncertainty makes it a robust choice for real-world applications, particularly in drug release studies where data noise is often unavoidable. Despite their computational expense, BPINNs offer a valuable enhancement over traditional PINNs, providing more reliable and interpretable results in the face of experimental data uncertainty.

\subsection*{PINN Predictive Abilities with Limited Experimental Data}

The evaluation of predictive models under noisy conditions revealed distinct trends across the film types, indicating varying model performances based on drug release complexity. As shown in Figure~\ref{fig:predict_error}, both Higuchi’s and Peppas’s models demonstrated limited effectiveness, with RMSE values being high, especially when trained on limited early time points and predicting later time points. For the 1D and 2D films, Fick’s model yielded RMSE values within the acceptable range only at higher \( n \) values, aligning with its suitability for diffusion-dominated processes. In Figure~\ref{fig:best_case}, the models with the lowest \( n \) where the PINN achieves an RMSE of less than 0.05 are shown. Figure~\ref{fig:predict_error} visually represents the PINN’s ability to accurately predict drug release with low error using early-stage data for training and later-stage data for testing.

\
The PINN consistently provided reliable predictions across all film types, achieving the best RMSE values below 0.05 for at lower \( n \) compared to the other models. In addition, it achieved the lowest RMSE compared to all other models. The lowest were \( n=9 \) for Flat, and \( n=11 \) for 1D winkled and 2D crumpled, as seen in Figure~\ref{fig:best_case}. These results show the PINN’s adaptability and accuracy, supporting its role as a robust predictive tool in drug delivery modeling, particularly under experimental constraints of limited data, as seen in Figure~\ref{fig:predict_error}. {For the flat film, the PINN achieves RMSE $<0.05$ with only $n=9$ points, which corresponds to 120 minutes (2 hours) of release data. In comparison, classical models require at least $n=12$–$13$ points (1–1.5 days of data) to reach similar accuracy, representing a savings of nearly a full day of experimental time. For the 1D and 2D films, the PINN reaches the same error threshold at $n=11$ (12 hours), whereas classical models again need $n=12$–$13$ (1–1.5 days). Thus, the PINN enables accurate long-term predictions 12–34 hours earlier, highlighting its value in accelerating drug release characterization. The superior performance of the PINN, which is able to model both  physical laws and experimental data highlights the potential for hybrid models to bridge the gap where classical approaches fall short, offering more reliable predictions in scenarios with sparse data.

\subsection*{Implications for Drug Delivery Modeling}
The findings from this study have significant implications for the future of drug delivery modeling, especially within the pharmaceutical and biotechnology industries. Classical models such as Higuchi’s model, Peppas’s model, and Fick’s Law have long served as the foundation for understanding and simulating drug release kinetics. These models, while powerful, operate under several idealized assumptions such as constant diffusivity and homogeneity of the drug matrix, which often fall short in representing complex, real-world scenarios \cite{siepmann2008mathematical, higuchi1961rate}. 

The introduction of PINNs into this landscape offers a substantial leap forward. Unlike classical models that rely strictly on predefined equations, PINNs combine the best of both worlds by integrating physical laws (like Fick’s Law) directly into the learning process while leveraging the flexibility of machine learning to fit real data more accurately. Our results reveal that the PINN not only replicates but also outperforms the predictive power of classical models, {achieving up to 40\% lower MAE and RMSE across all film types. Moreover, PINNs reach an RMSE below 0.05 with as few as 9–11 data points, whereas classical models require substantially more data to approach similar accuracy.}

This advantage of PINNs in achieving high predictive accuracy with fewer data points is particularly impactful. It represents a shift from purely equation-driven modeling to a more adaptive, data-integrated approach, reducing the reliance on extensive experimental data. With PINNs, we are no longer confined to the limitations of classical models, which often struggle to accommodate experimental noise, complex boundary conditions, and non-homogeneous materials in drug formulations. PINNs bring greater precision and robustness to drug delivery predictions, which can drastically reduce trial-and-error in the drug development pipeline, saving time and resources in the design of controlled-release systems. Furthermore, the implications for pharmaceutical applications are far-reaching. The consistent superiority of the PINN across different drug release profiles suggests that such a hybrid modeling approach could become a new standard for simulating drug kinetics, especially in personalized medicine, where drug release may vary across patient populations. By providing more accurate predictions with minimal data, PINNs can help optimize dosing regimens and formulation strategies, ultimately improving therapeutic outcomes and reducing side effects. This represents a promising advancement in data-driven approaches for drug delivery. Integrating physical laws with machine learning offers a foundation for next-generation pharmaceutical technologies aimed at enhancing therapeutic efficacy and enabling precise long-term drug release.

\begin{figure}[h!]
    \centering
    \includegraphics[width=0.8\textwidth]{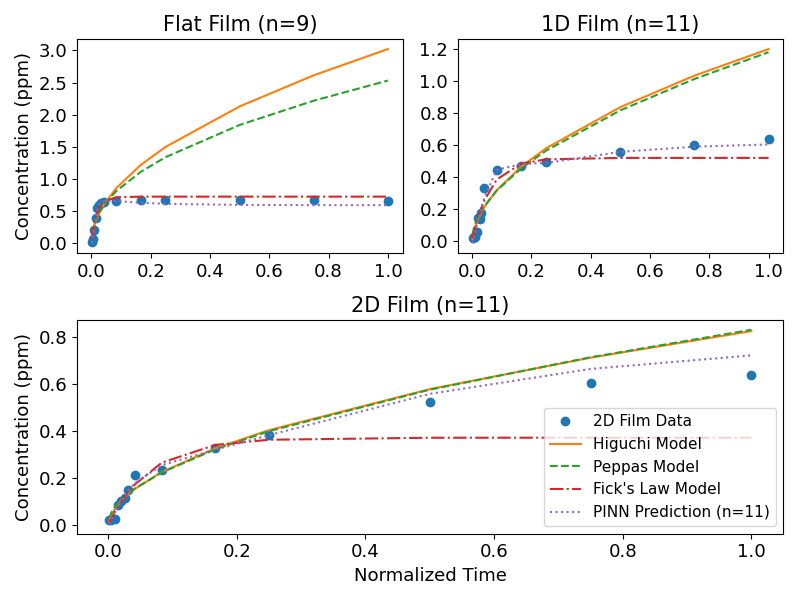}
    \caption{Comparison of PINN and classical models with experimental drug release data for flat, 1D, and 2D films at the minimum data points \( n \) achieving an RMSE below 0.05 for the PINN. (Top Left) Flat film with \( n = 9 \): The PINN model rapidly converges with fewer data points, accurately capturing the release profile. (Top Right) 1D film with \( n = 11 \): The PINN model maintains accuracy with reduced data, demonstrating a close fit to experimental data. (Bottom) 2D film with \( n = 11 \): PINN predictions show strong alignment with the data, reflecting the model's ability to generalize effectively across complex geometries. PINNs outperform classical models in terms of convergence speed and accuracy, particularly in capturing early-stage release dynamics with minimal data points. The total time (normalized time = 1) for all release curves is 48 hours. }
    \label{fig:best_case}
\end{figure}

\begin{figure}[h!]
    \centering
    \includegraphics[width=0.6\textwidth]{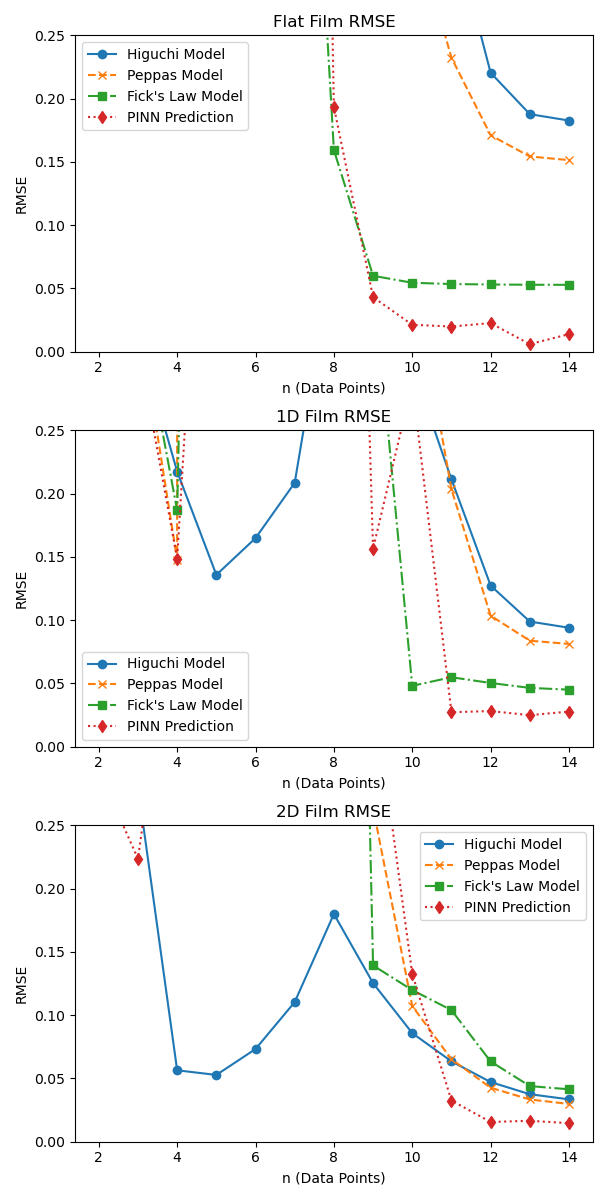}
     \caption{{RMSE values for classical models (Higuchi, Peppas, and Fick’s Law) compared with PINN predictions across Flat, 1D, and 2D films as a function of the number of data points $n$. For the flat film, the PINN achieves RMSE $<0.05$ at $n=9$ (corresponding to 120 minutes of release data), while classical models require $n=12$–$13$ (1440–2160 minutes), saving nearly a full day of experimental time. For the 1D and 2D films, the PINN reaches the same threshold at $n=11$ (720 minutes), compared to $n=12$–$13$ for classical models, saving 12–36 hours. These results highlight the PINN’s ability to use short-term release data to accurately predict long-term release, reducing the experimental burden.}}
    \label{fig:predict_error}
\end{figure}

\section*{Closing remarks}
This study demonstrated the effectiveness of PINNs and BPINNs in enhancing the accuracy of drug release modeling across different film types. By integrating Fick's Law with modern machine learning techniques, significant improvements in predictive performance were achieved. The PINN models consistently outperformed classical methods, as shown by lower MAE and RMSE, while also maintaining the interpretability of the underlying physical laws.

In particular, the ability of BPINNs to quantify uncertainty in noisy data makes them a powerful tool for handling experimental variability. This hybrid approach offers a more robust and adaptable solution, especially when dealing with complex geometries or non-ideal experimental conditions where classical models may fall short.

A key advantage of PINNs lies in their strong predictive capabilities, even when trained on minimal data points. The analysis in this study demonstrated that PINNs utilizing Fick's as physical law achieve RMSE values below the 0.05 threshold with as few as 9 data points in some cases, outperforming classical models, which require more data to reach similar accuracy. This predictive ability arises because the network is constrained by Fick’s diffusion law: once short-term data points help the PINN calibrate key parameters such as the diffusion coefficient, the governing PDE enforces physically consistent behavior over longer time horizons. In effect, the model leverages the physics prior to extrapolating release dynamics beyond the observed data. This efficiency in prediction with limited data highlights the potential for PINNs as a drug release predicting tool, particularly in experimental setups where limited data collection may be constrained. By learning from only a short-term release profile, the proposed PINN method builds a more accurate physics-constrained model of diffusion, which in turn enables reliable prediction of the long-term release behavior. This characteristic not only enhances the applicability of PINNs in early-stage drug formulation but also supports their use in personalized medicine, where patient-specific data is often scarce. In practice, this means fewer experiments are needed to characterize a delivery system, while still obtaining accurate long-term release predictions that classical models cannot provide. The ability to maintain accuracy with fewer data points can streamline the modeling process, making PINNs a valuable tool for accelerating drug development timelines.

These results highlight the broader potential of hybrid modeling techniques in drug delivery systems. The combination of physics-based laws and data-driven models provides a more accurate and flexible framework for simulating drug release, making it valuable for the pharmaceutical and biotech industries. This methodology can be further extended to optimize drug formulations, simulate novel drug delivery systems, and improve the accuracy of preclinical drug release behavior testing, thereby reducing costs and time associated with drug development. The incorporation of such techniques could lead to more reliable, efficient, and patient-specific therapies.

\section*{Data Availability}
The data used in this study and the code for implementing the PINN will be shared through GitHub. 

\section*{Acknowledgements}
We thank the Brown University School of Engineering and the Division of Applied Mathematics for their support in this research. We also thank Zachary Saleeba for his support during the initial part of the studyy.

\bibliographystyle{elsarticle-num}
\bibliography{references}

\end{document}